\documentclass[sigconf, nonacm]{acmart}
\AtBeginDocument{%
  \providecommand\BibTeX{{%
    \normalfont B\kern-0.5em{\scshape i\kern-0.25em b}\kern-0.8em\TeX}}}

\usepackage[]{svg}
\usepackage{lineno}

\begin{document}

\title{Inclusion in Assistive Haircare Robotics: \\ Practical and Ethical Considerations in Hair Manipulation}


\author{Uksang Yoo}
\email{uyoo@andrew.cmu.edu}
\affiliation{%
  \institution{Carnegie Mellon University Robotics Institute }
  \streetaddress{5000 Forbes Avenue}
  \city{Pittsburgh}
  \state{Pennsylvania}
  \country{USA}
  \postcode{15213}
}
\author{Nathaniel Dennler}
\email{dennler@usc.edu}
\affiliation{%
  \institution{University of Southern California Computer Science Department}
  \streetaddress{941 Bloom Walk}
  \city{Los Angeles}
  \state{California}
  \country{USA}
  \postcode{90089}
}

\author{Sarvesh Patil}
\email{sarveshp@andrew.cmu.edu}
\affiliation{%
  \institution{Carnegie Mellon University Robotics Institute }
  \streetaddress{5000 Forbes Avenue}
  \city{Pittsburgh}
  \state{Pennsylvania}
  \country{USA}
  \postcode{15213}
}
\author{Jean Oh}
\email{hyaejino@andrew.cmu.edu}
\affiliation{%
  \institution{Carnegie Mellon University Robotics Institute }
  \streetaddress{5000 Forbes Avenue}
  \city{Pittsburgh}
  \state{Pennsylvania}
  \country{USA}
  \postcode{15213}
}
\author{Jeffrey Ichnowski}
\email{jichnows@andrew.cmu.edu}
\affiliation{%
  \institution{Carnegie Mellon University Robotics Institute }
  \streetaddress{5000 Forbes Avenue}
  \city{Pittsburgh}
  \state{Pennsylvania}
  \country{USA}
  \postcode{15213}
}

\begin{abstract}
Robot haircare systems could provide a controlled and personalized environment that is respectful of an individual's sensitivities and may offer a comfortable experience. We argue that because of hair and hairstyles' often unique importance in defining and expressing an individual's identity, we should approach the development of assistive robot haircare systems carefully while considering various practical and ethical concerns and risks. In this work, we specifically list and discuss the consideration of hair type, expression of the individual's preferred identity, cost accessibility of the system, culturally-aware robot strategies, and the associated societal risks. Finally, we discuss the planned studies that will allow us to better understand and address the concerns and considerations we outlined in this work through interactions with both haircare experts and end-users. Through these practical and ethical considerations, this work seeks to systematically organize and provide guidance for the development of inclusive and ethical robot haircare systems.    
\end{abstract}



\maketitle
\section{Introduction}

Hair often occupies a crucial role in an individual's identity~\cite{batchelor_hair_2001,lashley_importance_2020}, and robot haircare systems hold promise for restoring independence and self-expression~\cite{dennler_design_2021}. For many aging individuals experiencing a decline in mobility, hair care has become an increasingly time-consuming and challenging daily task. At the same time, the significance of hair for self-esteem often grows with age~\cite{ward_if_2011}. Given that elder care and hospice facilities largely depend on volunteers for hair-care assistance~\cite{burbeck_understanding_2014}, there is a clear need for robotic systems to automate these tasks. Furthermore, cultural and sensory sensitivities can deter some individuals from visiting hair salons. For instance, neurodivergent people may experience sensory overload from the sounds, smells, and physical sensations of haircuts, making salon visits distressing~\cite{buckley2020teaching}. Robot haircare systems could provide a more controlled and personalized environment, addressing these sensitivities and offering a more comfortable experience.

 To address the haircare services gap, researchers have proposed deploying robot assistance for combing~\cite{dennler_design_2021, hughes_detangling_2021}. Assistive hair manipulation and styling present interesting challenges for roboticists such as trajectory planning based on hair flow estimation \cite{dennler_design_2021} and sensor-fusion for safe interaction with the head \cite{hughes_detangling_2021}. Addressing this year's theme of the Conference on Human Robot Interaction, ``HRI in the real world,'' we outline some practical and ethical considerations toward developing and transitioning assistive haircare robotic systems from controlled academic settings to the real world.

\section{Background}
In this section, we review works relevant to our discussion of robot haircare systems. Inclusive robot haircare systems should consider diverse physical hair types (2.1) as well as the users' identities (2.2). We also incorporate relevant literature on how identity has been considered in other areas of HRI research (2.3).

\subsection{Hair Types}
Proper categorization of hair's physical properties can help us develop and study robotic haircare strategies systematically. In the field of assistive feeding, the categorization of food items on the axes of their physical properties has helped to formulate robotic manipulation strategies systematically toward developing more robust robotic agents~\cite{bhattacharjee2019towards, feng2019robot}. By utilizing explicit categories of hair, we could similarly focus our efforts on addressing robotic haircare with diverse hair types. 

Early efforts to categorize hair types were built on problematic language and prejudice that conflated the construct of race with hair properties~\cite{eddy1938hair}. Such works largely used these hair property categories to define and distinguish racial groups, often with categories that are rejected today. 

Since then, researchers have produced various categorization schemes for hair textures and properties~\cite{loussouarn2007worldwide, moody2022impact}. Many of these contemporary approaches to understanding different hair types largely attempt to decouple the notion of race from the physical properties of hair such as its length, cross-sectional thickness, and curliness. Previous works have explicitly recognized the problematic history of hair categorization and actively addressed them in trying to create hair property categorization schemes that are not grounded in racial identities~\cite{moody2022impact}. However, some of the physical properties of hair used in these categorization schemes are still often referenced in racialized contexts in literature~\cite{leerunyakul2020asian, moody2022impact}. Additionally, the process of developing hair categorization schemes often attempts to collect a representative sample from the global population~\cite{loussouarn2007worldwide}, which may reflect today's notion of race and ethnic identity. 

\subsection{Identity Expressions with Hair}
An inclusive robot haircare system should not only be able to handle diverse types of hair but also enable the users to express their preferred identities reflected in diverse hairstyles. The way that people style hair has long played a role in defining and expressing national, ethnic, religious, and racial identities~\cite{olivelle200815,barritt2021sexual,robins1999hair, son2010study}.  Individuals may also choose to express their sexual and gender identities through hairstyles~\cite{reddy2018cut,clarke2013navigating}. Importantly, these identities may intersect and be expressed through external appearance and hairstyle~\cite{cerezo2020identity}.

An individual's hairstyle also significantly affects what others presume about the individual's identity, even controlling for other visible features~\cite{sims2020doing}. While a component of such social interaction with hairstyle can be an individual's avenue for identity expression~\cite{reddy2018cut}, it can also be a source of prejudice and oppression~\cite{chaves2021hair}. For instance, there is an active call to protect the freedom to express one's identity through hairstyle in the United States~\cite{halbert2021hair}. Such struggles provide social context for the development of an inclusive robotic haircare system that can not only handle diverse hair types but also enable users to express diverse preferred identities through hairstyles.
\subsection{Identity-based Personalization}
Researchers in both Human-Computer Interaction (HCI) and HRI have studied various approaches to identity-aware and identity-based personalization of user experience. Such approaches have sometimes called to explicitly consider the user's race~\cite{liao2020racial} or gender~\cite{cruz2017learning,saggese2019miviabot}. Some researchers argue that explicitly considering the user's identity to personalize robotic interaction experiences can result in improved efficacy and enable effective consideration of cultural differences~\cite{gasteiger2023factors}. Linking identity to a user can affirm their identity, reduce identity erasure, promote equitable resource allocation, and reduce the chance that a user may experience content that causes dysphoria ~\cite{dennler2023bound}. Additionally, broader frameworks such as Design Justice argue that we should explicitly consider groups of people that benefit from and experience the burden of systems ~\cite{costanza2020design}. 

There has also been a criticism of identity-based personalization because explicitly classifying the gender or race of the user could lead to reinforcing harmful norms or cause harm by misidentifying the user's identity~\cite{williams2023eye}. Interviews with developers and users of AI systems have qualified that systems collecting explicit representations of identity should be optional, mutable, and revocable at any time ~\cite{dennler2023bound}. Whether we explicitly or implicitly consider the user's identity is a naturally relevant topic for the development of robot haircare systems because of hairstyle's link to identity expression.

\section{Considerations}
\begin{figure}[t!]

  \includegraphics[width=1.0\columnwidth]{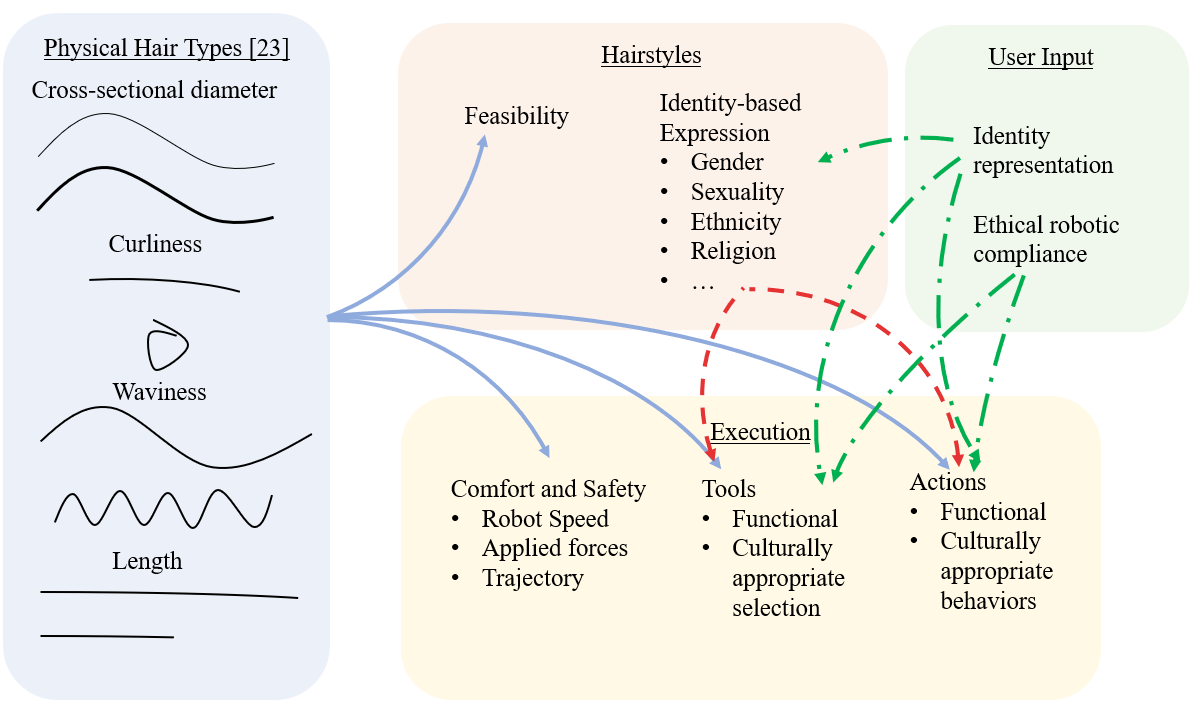}
  \caption{Functional diagram of the practical and ethical considerations for robot haircare system development. The physical hair type directly determines the space of feasible hairstyles. Additionally, it affects what robot speeds, applied forces and trajectories are physically comfortable for the users. It also affects what tools can be used and what robot actions are effective. Identity-based expression through hairstyles partially determines the culturally appropriate tools and actions the robot can perform. The user input of identity directly relates to the consideration of hairstyles that reflect the user's expressed identity. The user's identity input and which user commands the system determines are ethical to comply with relate to the concerns of culturally appropriate tools and actions that the robot should use. }
  \label{figure:diagram}
  \vspace{-10pt}
\end{figure}

In this section, we list and briefly outline the considerations we believe should be incorporated into the development of an inclusive and ethical robot haircare system as outlined in Fig. \ref{figure:diagram}. Toward inclusion, we should consider how we can ensure that the developed robot haircare system can address diverse hair types (3.1) and allow comfortable expression of the user's identity (3.2). Additionally, we must consider the possible barriers to adoption rooted in the system's associated costs and economic disparities (3.3). Through the system development, we should also consider the potential risks of some design choices such as user identity representation, and minimize harm (3.4).
\subsection{Hair Types}
Categorization of the physical properties of an individual's hair can provide guidance and structure to systematically develop robotic haircare strategies that can be functional with diverse hair types. Crucially, we can make an effort to not explicitly consider race at the perception level by utilizing categorization schemes of hair types grounded directly on individual hair's physical properties as opposed to the user's perceived or self-identified ethnicity. A well-grounded categorization of the user's hair is a necessary step to develop a robot haircare system that can effectively manipulate diverse types of hair. For example, a manipulation strategy to comb straight hair has the potential to permanently damage hair structures in curly hair. Employing tools like counterfactual reasoning helps eliminate biases from learning-based systems as shown in \cite{counterfactual_1, counterfactual_2}, enabling the development of effective characterization schemes. These effective categorization schemes for hair types can help the system designers systematically discover gaps in the robotic system's covered serviceable population. Additionally, the categorization of hair types also informs admissible actions, relevant transition models, and manipulation strategies that the assistive haircare robot can perform. The hair type also defines the space of feasible hairstyles that the user can hope to achieve with the system.

\subsection{Identity-based Expression and Preferences}
When exploring the space of feasible hairstyles defined by the hair type, we should explicitly consider identity-expressive hairstyles to ensure that they are appropriately represented in the design of the system. In addition to the hairstyle that the robot haircare system should be able to address, it should also use culturally appropriate skills to style the hair. Using culturally appropriate skills such as making dreadlocks or rishi knots and using tools such as hair picks may not only contribute to added physical comfort (e.g., by not pulling on the hair excessively), it may contribute to the robot system behaving more closely aligned to the user's expectations.


\subsection{Cost}
Even if the robot haircare system is functionally inclusive and robust, associated costs for the purchase and operation may in practice prohibit access to many marginalized communities and regions. Analogously, researchers have highlighted challenges faced by low- and middle-income countries in adopting robots for stroke rehabilitation~\cite{demofonti2021affordable}. Additionally, many state-of-the-art robotic healthcare and assistive technologies have largely been made available only to the wealthier socioeconomic groups even in high-income countries such as the United States~\cite{tatarian2023socioeconomic}.

Without the consideration of cost, haircare robot systems and technologies may work to widen the socioeconomic disparity in various areas of life such as elderly and disability care. Taking inspiration from related efforts in medical and rehabilitation robots~\cite{demofonti2021affordable,sarkar2019low,gaardsmoe2020development}, there should be an effort to develop robot haircare systems that are affordable, robust to different environments, locally repairable, and easily operable.

\subsection{Risks and Concerns}
Some researchers have recently raised concerns over the ethical risks associated with identity-based personalization of robot agents~\cite{williams2023eye}.  A primary concern raised by researchers is that explicitly identifying identities may work to inadvertently propagate problematic and prejudiced notions of race, gender, and other dimensions of identity. 

For robot haircare, some degree of identity-related classification is required as the robot must be able to perceive and understand various physical properties of the hair and reason about the user's preferences. Such perceived properties may reflect the user's identity as proxy identifiers. For instance, certain ranges of hair thickness, straightness, and color may cumulatively be associated with individuals of a certain race~\cite{leerunyakul2020asian}. As such, much of the considerations made in this section assume that robot haircare systems will at least implicitly reason about the user's identity. However, we could minimize the risk of harm by avoiding explicit consideration of the user's identity and taking into consideration the user's self-identified preferences for the desired style. 

Robust preference communication between the human and the robot agents should be a component of an inclusive and useful robotic haircare system. In some HRI applications, however, complete compliance of the robotic agent to human commands may result in problematic results~\cite{briggs2022and}. For instance, particularly for haircare, the developed robot system may further enable offensive human behaviors such as cultural appropriation of hairstyles ~\cite{chaves2021hair} in the United States. Appropriation of hairstyles and other harmful or offensive behaviors should be explicitly considered in the development of a robotic haircare system that can recognize and appropriately refuse user requests that may do social harm or offend norms. 

\section{Planned Study}

In order to design a hair ontology to define research directions for assistive robotic haircare, we aim to gather information from a variety of stakeholders in the system. We are interested in both experts in haircare as well as actual end-users of the system.

\subsection{Data Collection}
We aim to interview three types of participants: hair-care professionals, caretakers, and participants with limited mobility. These participants will be recruited from populations in the surrounding community through flyers, email lists, and word-of-mouth. Participants will engage in a 30-minute semi-structured online interviews. 

For hair-care professionals and caretakers, the interview will focus on the different considerations needed for manipulating hair on people of varying identities along racial, ethnic, gender, and sexual orientation axes. This interviews will investigate what strategies a robot may need to take to work well with varying hair types. These participants will discuss to what extent different types of of hair are similar and in what ways they differ. These participants may also speak about the various impacts of haircare beyond functional assistance, which will inform how an assistive haircare system should be designed.

For participants affected by limited mobility, the focus of the interview will revolve around how participants expect to interact with a system that provides assistance with haircare. We also aim to gather any reservations, concerns, or anticipated negative effects of using a system for assistive haircare. By interviewing the anticipated end-users we can develop an understanding for the ways the robots can maximally meet the needs of this specific population, incorporating any potential harms of the system directly into the design process.

We aim to recruit approximately 20 participants from each category of participant to interview, for a total of 60 participants. The recruitment will officially end when the study reaches theoretical saturation as defined by Saunders et al. \cite{saunders2018saturation}. Given the limited size of the dataset due to location constraints, we plan to augment the semi-structured interviews with data collected from videos posted online. This can allow us to incorporate a wider variety of identities that may not be recruited to partake in the semi-structured interviews.

\subsection{Proposed Analysis}
After interviewing the participants, we aim to analyze the findings through a iterative inductive thematic analysis of the interview transcripts and collected haircare videos. The procedure will follow the following steps, conducted across a research team consisting of researchers of various identities: (1) all interviews transcripts and haircare videos will be coded with topics of discussion, (2) researchers will independently compile these codes into different themes, (3) researchers will meet as a team to discuss the themes that emerged, (4) concepts will be grouped into hierarchical themes, and (5) steps 2-4 will be iteratively repeated until the team reaches a consensus on what themes are important for assistive robotic haircare.

\subsection{Mitigating Risks and Biases}
In order to mitigate the risks associated with collecting data from vulnerable populations, we will take several considerations. First, we will submit the proposed protocol to an IRB to verify that that study follows ethical guidelines. Participants will be directly and clearly informed of what is expected of them and the overall procedure of the study. 

To reduce biases in the data itself, the interview population will be sampled to include identities that span several intersecting axes. In addition, the interviews will be conducted with interviewers of varying identities. Ideally, interviewers and interviewees will have some shared experiences to establish and build rapport between the researcher and participant, as is common in ethnographic research \cite{glesne1989rapport}. Having established rapport with the interviewee can elicit more genuine responses to research questions and mitigate biases that may be introduced through demand characteristics, perceived authority, and other power dynamics.

To reduce biases in the analysis of the data, we will perform a thematic analysis with several researchers of varying identities. The identities of the analysts will be critically examined to determine how identities shape the outcomes of the analysis, and the reflections will be summarized in a positionality statement.

\section{Conclusion and Future Work}
We outlined some of the practical and ethical considerations that should be made in the development of robot haircare systems. Primarily, we argue that we should not only consider the hair type of the user but also an expression of the user's preferred identities. Additionally, we outline various risks and concerns associated with the development of a robot haircare system and propose mitigation approaches to minimize potential harm. As an extension of this work, we plan to construct a robot haircare manipulation strategy taxonomy with the considerations outlined in this work. 


\bibliographystyle{ACM-Reference-Format}
\bibliography{ref}


\begin{thebibliography}{39}


\ifx \showCODEN    \undefined \def \showCODEN     #1{\unskip}     \fi
\ifx \showDOI      \undefined \def \showDOI       #1{#1}\fi
\ifx \showISBNx    \undefined \def \showISBNx     #1{\unskip}     \fi
\ifx \showISBNxiii \undefined \def \showISBNxiii  #1{\unskip}     \fi
\ifx \showISSN     \undefined \def \showISSN      #1{\unskip}     \fi
\ifx \showLCCN     \undefined \def \showLCCN      #1{\unskip}     \fi
\ifx \shownote     \undefined \def \shownote      #1{#1}          \fi
\ifx \showarticletitle \undefined \def \showarticletitle #1{#1}   \fi
\ifx \showURL      \undefined \def \showURL       {\relax}        \fi
\providecommand\bibfield[2]{#2}
\providecommand\bibinfo[2]{#2}
\providecommand\natexlab[1]{#1}
\providecommand\showeprint[2][]{arXiv:#2}

\bibitem[Barritt(2021)]%
        {barritt2021sexual}
\bibfield{author}{\bibinfo{person}{Julian~Ash Barritt}.} \bibinfo{year}{2021}\natexlab{}.
\newblock \showarticletitle{Sexual Orientation, Gender, \& Self-Styling: An Exploration of Visual Identity-Signaling}.
\newblock  (\bibinfo{year}{2021}).
\newblock


\bibitem[Batchelor(2001)]%
        {batchelor_hair_2001}
\bibfield{author}{\bibinfo{person}{D. Batchelor}.} \bibinfo{year}{2001}\natexlab{}.
\newblock \showarticletitle{Hair and cancer chemotherapy: consequences and nursing care - a literature study: \textit{{European} {Journal} of {Cancer} {Care}}}.
\newblock \bibinfo{journal}{\emph{European Journal of Cancer Care}} \bibinfo{volume}{10}, \bibinfo{number}{3} (\bibinfo{date}{Sept.} \bibinfo{year}{2001}), \bibinfo{pages}{147--163}.
\newblock
\showISSN{09615423}
\urldef\tempurl%
\url{https://doi.org/10.1046/j.1365-2354.2001.00272.x}
\showDOI{\tempurl}


\bibitem[Bhattacharjee et~al\mbox{.}(2019)]%
        {bhattacharjee2019towards}
\bibfield{author}{\bibinfo{person}{Tapomayukh Bhattacharjee}, \bibinfo{person}{Gilwoo Lee}, \bibinfo{person}{Hanjun Song}, {and} \bibinfo{person}{Siddhartha~S Srinivasa}.} \bibinfo{year}{2019}\natexlab{}.
\newblock \showarticletitle{Towards robotic feeding: Role of haptics in fork-based food manipulation}.
\newblock \bibinfo{journal}{\emph{IEEE Robotics and Automation Letters}} \bibinfo{volume}{4}, \bibinfo{number}{2} (\bibinfo{year}{2019}), \bibinfo{pages}{1485--1492}.
\newblock


\bibitem[Briggs et~al\mbox{.}(2022)]%
        {briggs2022and}
\bibfield{author}{\bibinfo{person}{Gordon Briggs}, \bibinfo{person}{Tom Williams}, \bibinfo{person}{Ryan~Blake Jackson}, {and} \bibinfo{person}{Matthias Scheutz}.} \bibinfo{year}{2022}\natexlab{}.
\newblock \showarticletitle{Why and how robots should say ‘no’}.
\newblock \bibinfo{journal}{\emph{International Journal of Social Robotics}} \bibinfo{volume}{14}, \bibinfo{number}{2} (\bibinfo{year}{2022}), \bibinfo{pages}{323--339}.
\newblock


\bibitem[Buckley et~al\mbox{.}(2020)]%
        {buckley2020teaching}
\bibfield{author}{\bibinfo{person}{Jessica Buckley}, \bibinfo{person}{James~K Luiselli}, \bibinfo{person}{Jill~M Harper}, {and} \bibinfo{person}{Andrew Shlesinger}.} \bibinfo{year}{2020}\natexlab{}.
\newblock \showarticletitle{Teaching students with autism spectrum disorder to tolerate haircutting}.
\newblock \bibinfo{journal}{\emph{Journal of Applied Behavior Analysis}} \bibinfo{volume}{53}, \bibinfo{number}{4} (\bibinfo{year}{2020}), \bibinfo{pages}{2081--2089}.
\newblock


\bibitem[Burbeck et~al\mbox{.}(2014)]%
        {burbeck_understanding_2014}
\bibfield{author}{\bibinfo{person}{Rachel Burbeck}, \bibinfo{person}{Bridget Candy}, \bibinfo{person}{Joe Low}, {and} \bibinfo{person}{Rebecca Rees}.} \bibinfo{year}{2014}\natexlab{}.
\newblock \showarticletitle{Understanding the role of the volunteer in specialist palliative care: a systematic review and thematic synthesis of qualitative studies}.
\newblock \bibinfo{journal}{\emph{BMC Palliative Care}} \bibinfo{volume}{13}, \bibinfo{number}{1} (\bibinfo{date}{Dec.} \bibinfo{year}{2014}), \bibinfo{pages}{3}.
\newblock
\showISSN{1472-684X}
\urldef\tempurl%
\url{https://doi.org/10.1186/1472-684X-13-3}
\showDOI{\tempurl}


\bibitem[Cerezo et~al\mbox{.}(2020)]%
        {cerezo2020identity}
\bibfield{author}{\bibinfo{person}{Alison Cerezo}, \bibinfo{person}{Mariah Cummings}, \bibinfo{person}{Meredith Holmes}, {and} \bibinfo{person}{Chelsey Williams}.} \bibinfo{year}{2020}\natexlab{}.
\newblock \showarticletitle{Identity as resistance: Identity formation at the intersection of race, gender identity, and sexual orientation}.
\newblock \bibinfo{journal}{\emph{Psychology of women quarterly}} \bibinfo{volume}{44}, \bibinfo{number}{1} (\bibinfo{year}{2020}), \bibinfo{pages}{67--83}.
\newblock


\bibitem[Chaves and Bacharach(2021)]%
        {chaves2021hair}
\bibfield{author}{\bibinfo{person}{Andrea~Mej{\'\i}a Chaves} {and} \bibinfo{person}{Sondra Bacharach}.} \bibinfo{year}{2021}\natexlab{}.
\newblock \showarticletitle{Hair oppression and appropriation}.
\newblock \bibinfo{journal}{\emph{The British Journal of Aesthetics}} \bibinfo{volume}{61}, \bibinfo{number}{3} (\bibinfo{year}{2021}), \bibinfo{pages}{335--352}.
\newblock


\bibitem[Clarke and Spence(2013)]%
        {clarke2013navigating}
\bibfield{author}{\bibinfo{person}{Victoria Clarke} {and} \bibinfo{person}{Katherine Spence}.} \bibinfo{year}{2013}\natexlab{}.
\newblock \showarticletitle{‘I am who I am’? Navigating norms and the importance of authenticity in lesbian and bisexual women's accounts of their appearance practices}.
\newblock \bibinfo{journal}{\emph{Psychology \& Sexuality}} \bibinfo{volume}{4}, \bibinfo{number}{1} (\bibinfo{year}{2013}), \bibinfo{pages}{25--33}.
\newblock


\bibitem[Costanza-Chock(2020)]%
        {costanza2020design}
\bibfield{author}{\bibinfo{person}{Sasha Costanza-Chock}.} \bibinfo{year}{2020}\natexlab{}.
\newblock \bibinfo{booktitle}{\emph{Design justice: Community-led practices to build the worlds we need}}.
\newblock \bibinfo{publisher}{The MIT Press}.
\newblock


\bibitem[Cruz-Maya and Tapus(2017)]%
        {cruz2017learning}
\bibfield{author}{\bibinfo{person}{Arturo Cruz-Maya} {and} \bibinfo{person}{Adriana Tapus}.} \bibinfo{year}{2017}\natexlab{}.
\newblock \showarticletitle{Learning users' and personality-gender preferences in close human-robot interaction}. In \bibinfo{booktitle}{\emph{2017 26th IEEE International Symposium on Robot and Human Interactive Communication (RO-MAN)}}. IEEE, \bibinfo{pages}{791--798}.
\newblock


\bibitem[Demofonti et~al\mbox{.}(2021)]%
        {demofonti2021affordable}
\bibfield{author}{\bibinfo{person}{Andrea Demofonti}, \bibinfo{person}{Giorgio Carpino}, \bibinfo{person}{Loredana Zollo}, {and} \bibinfo{person}{Michelle~J Johnson}.} \bibinfo{year}{2021}\natexlab{}.
\newblock \showarticletitle{Affordable robotics for upper limb stroke rehabilitation in developing countries: a systematic review}.
\newblock \bibinfo{journal}{\emph{IEEE Transactions on Medical Robotics and Bionics}} \bibinfo{volume}{3}, \bibinfo{number}{1} (\bibinfo{year}{2021}), \bibinfo{pages}{11--20}.
\newblock


\bibitem[Dennler et~al\mbox{.}(2023)]%
        {dennler2023bound}
\bibfield{author}{\bibinfo{person}{Nathan Dennler}, \bibinfo{person}{Anaelia Ovalle}, \bibinfo{person}{Ashwin Singh}, \bibinfo{person}{Luca Soldaini}, \bibinfo{person}{Arjun Subramonian}, \bibinfo{person}{Huy Tu}, \bibinfo{person}{William Agnew}, \bibinfo{person}{Avijit Ghosh}, \bibinfo{person}{Kyra Yee}, \bibinfo{person}{Irene~Font Peradejordi}, {et~al\mbox{.}}} \bibinfo{year}{2023}\natexlab{}.
\newblock \showarticletitle{Bound by the Bounty: Collaboratively Shaping Evaluation Processes for Queer AI Harms}. In \bibinfo{booktitle}{\emph{Proceedings of the 2023 AAAI/ACM Conference on AI, Ethics, and Society}}. \bibinfo{pages}{375--386}.
\newblock


\bibitem[Dennler et~al\mbox{.}(2021)]%
        {dennler_design_2021}
\bibfield{author}{\bibinfo{person}{Nathaniel Dennler}, \bibinfo{person}{Eura Shin}, \bibinfo{person}{Maja Mataric}, {and} \bibinfo{person}{Stefanos Nikolaidis}.} \bibinfo{year}{2021}\natexlab{}.
\newblock \showarticletitle{Design and {Evaluation} of a {Hair} {Combing} {System} {Using} a {General}-{Purpose} {Robotic} {Arm}}. In \bibinfo{booktitle}{\emph{2021 {IEEE}/{RSJ} {International} {Conference} on {Intelligent} {Robots} and {Systems} ({IROS})}}. \bibinfo{publisher}{IEEE}, \bibinfo{address}{Prague, Czech Republic}, \bibinfo{pages}{3739--3746}.
\newblock
\showISBNx{978-1-66541-714-3}
\urldef\tempurl%
\url{https://doi.org/10.1109/IROS51168.2021.9636768}
\showDOI{\tempurl}


\bibitem[Eddy(1938)]%
        {eddy1938hair}
\bibfield{author}{\bibinfo{person}{MW Eddy}.} \bibinfo{year}{1938}\natexlab{}.
\newblock \showarticletitle{Hair classification}. In \bibinfo{booktitle}{\emph{Proceedings of the Pennsylvania Academy of Science}}, Vol.~\bibinfo{volume}{12}. JSTOR, \bibinfo{pages}{19--26}.
\newblock


\bibitem[Feng et~al\mbox{.}(2019)]%
        {feng2019robot}
\bibfield{author}{\bibinfo{person}{Ryan Feng}, \bibinfo{person}{Youngsun Kim}, \bibinfo{person}{Gilwoo Lee}, \bibinfo{person}{Ethan~K Gordon}, \bibinfo{person}{Matt Schmittle}, \bibinfo{person}{Shivaum Kumar}, \bibinfo{person}{Tapomayukh Bhattacharjee}, {and} \bibinfo{person}{Siddhartha~S Srinivasa}.} \bibinfo{year}{2019}\natexlab{}.
\newblock \showarticletitle{Robot-assisted feeding: Generalizing skewering strategies across food items on a plate}. In \bibinfo{booktitle}{\emph{The International Symposium of Robotics Research}}. Springer, \bibinfo{pages}{427--442}.
\newblock


\bibitem[Gaardsmoe et~al\mbox{.}(2020)]%
        {gaardsmoe2020development}
\bibfield{author}{\bibinfo{person}{Samuel Gaardsmoe}, \bibinfo{person}{Maria Ovando}, \bibinfo{person}{Kevin Bui}, {and} \bibinfo{person}{Michelle~J Johnson}.} \bibinfo{year}{2020}\natexlab{}.
\newblock \showarticletitle{Development of a low-cost balance assessment system for use in an affordable robot gym in low and middle income countries}. In \bibinfo{booktitle}{\emph{2020 IEEE 11th Latin American Symposium on Circuits \& Systems (LASCAS)}}. IEEE, \bibinfo{pages}{1--6}.
\newblock


\bibitem[Gasteiger et~al\mbox{.}(2023)]%
        {gasteiger2023factors}
\bibfield{author}{\bibinfo{person}{Norina Gasteiger}, \bibinfo{person}{Mehdi Hellou}, {and} \bibinfo{person}{Ho~Seok Ahn}.} \bibinfo{year}{2023}\natexlab{}.
\newblock \showarticletitle{Factors for personalization and localization to optimize human--robot interaction: A literature review}.
\newblock \bibinfo{journal}{\emph{International Journal of Social Robotics}} \bibinfo{volume}{15}, \bibinfo{number}{4} (\bibinfo{year}{2023}), \bibinfo{pages}{689--701}.
\newblock


\bibitem[Glesne(1989)]%
        {glesne1989rapport}
\bibfield{author}{\bibinfo{person}{Corrine Glesne}.} \bibinfo{year}{1989}\natexlab{}.
\newblock \showarticletitle{Rapport and friendship in ethnographic research}.
\newblock \bibinfo{journal}{\emph{Internation Journal of Qualitative Studies in Education}} \bibinfo{volume}{2}, \bibinfo{number}{1} (\bibinfo{year}{1989}), \bibinfo{pages}{45--54}.
\newblock


\bibitem[Halbert(2021)]%
        {halbert2021hair}
\bibfield{author}{\bibinfo{person}{Alexandra Halbert}.} \bibinfo{year}{2021}\natexlab{}.
\newblock \showarticletitle{Hair Goes Nothing: Proposing the Uniform Enactment of the Crown Act Across the United States}.
\newblock \bibinfo{journal}{\emph{Journal of Race, Gender, and Ethnicity}} \bibinfo{volume}{10}, \bibinfo{number}{1} (\bibinfo{year}{2021}), \bibinfo{pages}{11}.
\newblock


\bibitem[Hughes et~al\mbox{.}(2021)]%
        {hughes_detangling_2021}
\bibfield{author}{\bibinfo{person}{Josie Hughes}, \bibinfo{person}{Thomas Plumb-Reyes}, \bibinfo{person}{Nicholas Charles}, \bibinfo{person}{L. Mahadevan}, {and} \bibinfo{person}{Daniela Rus}.} \bibinfo{year}{2021}\natexlab{}.
\newblock \showarticletitle{Detangling hair using feedback-driven robotic brushing.}. In \bibinfo{booktitle}{\emph{4th {IEEE} {International} {Conference} on {Soft} {Robotics}, {RoboSoft} 2021, {New} {Haven}, {CT}, {USA}, {April} 12-16, 2021}}. \bibinfo{pages}{487--494}.
\newblock
\urldef\tempurl%
\url{https://doi.org/10.1109/ROBOSOFT51838.2021.9479221}
\showDOI{\tempurl}


\bibitem[Lashley(2020)]%
        {lashley_importance_2020}
\bibfield{author}{\bibinfo{person}{Myrna Lashley}.} \bibinfo{year}{2020}\natexlab{}.
\newblock \showarticletitle{The importance of hair in the identity of {Black} people}.
\newblock \bibinfo{journal}{\emph{Nouvelles pratiques sociales}} \bibinfo{volume}{31}, \bibinfo{number}{2} (\bibinfo{year}{2020}), \bibinfo{pages}{206--227}.
\newblock
\showISSN{1703-9312}
\urldef\tempurl%
\url{https://doi.org/10.7202/1076652ar}
\showDOI{\tempurl}
\newblock
\shownote{Publisher: Université du Québec à Montréal}.


\bibitem[Leerunyakul and Suchonwanit(2020)]%
        {leerunyakul2020asian}
\bibfield{author}{\bibinfo{person}{Kanchana Leerunyakul} {and} \bibinfo{person}{Poonkiat Suchonwanit}.} \bibinfo{year}{2020}\natexlab{}.
\newblock \showarticletitle{Asian hair: a review of structures, properties, and distinctive disorders}.
\newblock \bibinfo{journal}{\emph{Clinical, Cosmetic and Investigational Dermatology}} (\bibinfo{year}{2020}), \bibinfo{pages}{309--318}.
\newblock


\bibitem[Liao and He(2020)]%
        {liao2020racial}
\bibfield{author}{\bibinfo{person}{Yuting Liao} {and} \bibinfo{person}{Jiangen He}.} \bibinfo{year}{2020}\natexlab{}.
\newblock \showarticletitle{Racial mirroring effects on human-agent interaction in psychotherapeutic conversations}. In \bibinfo{booktitle}{\emph{Proceedings of the 25th international conference on intelligent user interfaces}}. \bibinfo{pages}{430--442}.
\newblock


\bibitem[Loussouarn et~al\mbox{.}(2007)]%
        {loussouarn2007worldwide}
\bibfield{author}{\bibinfo{person}{Genevi{\`e}ve Loussouarn}, \bibinfo{person}{Anne-Lise Garcel}, \bibinfo{person}{Isabelle Lozano}, \bibinfo{person}{Catherine Collaudin}, \bibinfo{person}{Crystal Porter}, \bibinfo{person}{S{\'e}gol{\`e}ne Panhard}, \bibinfo{person}{Didier Saint-L{\'e}ger}, {and} \bibinfo{person}{Roland De~La~Mettrie}.} \bibinfo{year}{2007}\natexlab{}.
\newblock \showarticletitle{Worldwide diversity of hair curliness: a new method of assessment}.
\newblock \bibinfo{journal}{\emph{International journal of dermatology}}  \bibinfo{volume}{46} (\bibinfo{year}{2007}), \bibinfo{pages}{2--6}.
\newblock


\bibitem[Moody et~al\mbox{.}(2022)]%
        {moody2022impact}
\bibfield{author}{\bibinfo{person}{Shannin~N Moody}, \bibinfo{person}{Lotte van Dammen}, \bibinfo{person}{Wen Wang}, \bibinfo{person}{Kimberly~A Greder}, \bibinfo{person}{Jenae~M Neiderhiser}, \bibinfo{person}{Patience~A Afulani}, \bibinfo{person}{Auriel Willette}, {and} \bibinfo{person}{Elizabeth~A Shirtcliff}.} \bibinfo{year}{2022}\natexlab{}.
\newblock \showarticletitle{Impact of hair type, hair sample weight, external hair exposures, and race on cumulative hair cortisol}.
\newblock \bibinfo{journal}{\emph{Psychoneuroendocrinology}}  \bibinfo{volume}{142} (\bibinfo{year}{2022}), \bibinfo{pages}{105805}.
\newblock


\bibitem[Olivelle(2008)]%
        {olivelle200815}
\bibfield{author}{\bibinfo{person}{Patrick Olivelle}.} \bibinfo{year}{2008}\natexlab{}.
\newblock \showarticletitle{15. Hair and Society: Social Significance of Hair in South Asian Traditions}.
\newblock \bibinfo{journal}{\emph{15. Hair and Society}} (\bibinfo{year}{2008}), \bibinfo{pages}{1000--1030}.
\newblock


\bibitem[Reddy-Best(2018)]%
        {reddy2018cut}
\bibfield{author}{\bibinfo{person}{Kelly Reddy-Best}.} \bibinfo{year}{2018}\natexlab{}.
\newblock \showarticletitle{‘I cut it [her hair] real short right after I got the job’: Queer coding during the interview for LGBTQ+ women}.
\newblock \bibinfo{journal}{\emph{Fashion, Style \& Popular Culture}} \bibinfo{volume}{5}, \bibinfo{number}{2} (\bibinfo{year}{2018}), \bibinfo{pages}{221--234}.
\newblock


\bibitem[Robins(1999)]%
        {robins1999hair}
\bibfield{author}{\bibinfo{person}{Gay Robins}.} \bibinfo{year}{1999}\natexlab{}.
\newblock \showarticletitle{Hair and the Construction of Identity in Ancient Egypt, c. 1480-1350 BC}.
\newblock \bibinfo{journal}{\emph{Journal of the American Research Center in Egypt}}  \bibinfo{volume}{36} (\bibinfo{year}{1999}), \bibinfo{pages}{55--69}.
\newblock


\bibitem[Saggese et~al\mbox{.}(2019)]%
        {saggese2019miviabot}
\bibfield{author}{\bibinfo{person}{Alessia Saggese}, \bibinfo{person}{Mario Vento}, {and} \bibinfo{person}{Vincenzo Vigilante}.} \bibinfo{year}{2019}\natexlab{}.
\newblock \showarticletitle{MIVIABot: a cognitive robot for smart museum}. In \bibinfo{booktitle}{\emph{Computer Analysis of Images and Patterns: 18th International Conference, CAIP 2019, Salerno, Italy, September 3--5, 2019, Proceedings, Part I 18}}. Springer, \bibinfo{pages}{15--25}.
\newblock


\bibitem[Sarkar et~al\mbox{.}(2019)]%
        {sarkar2019low}
\bibfield{author}{\bibinfo{person}{Dibya~Prokash Sarkar}, \bibinfo{person}{Md~Fahim Farden}, \bibinfo{person}{Md~Atiqul Islam}, \bibinfo{person}{Rahat~Jahangir Rony}, {and} \bibinfo{person}{Tamanna Motahar}.} \bibinfo{year}{2019}\natexlab{}.
\newblock \showarticletitle{A low-cost healthcare bot for elderly people}. In \bibinfo{booktitle}{\emph{2019 Joint 8th International Conference on Informatics, Electronics \& Vision (ICIEV) and 2019 3rd International Conference on Imaging, Vision \& Pattern Recognition (icIVPR)}}. IEEE, \bibinfo{pages}{18--23}.
\newblock


\bibitem[Saunders et~al\mbox{.}(2018)]%
        {saunders2018saturation}
\bibfield{author}{\bibinfo{person}{Benjamin Saunders}, \bibinfo{person}{Julius Sim}, \bibinfo{person}{Tom Kingstone}, \bibinfo{person}{Shula Baker}, \bibinfo{person}{Jackie Waterfield}, \bibinfo{person}{Bernadette Bartlam}, \bibinfo{person}{Heather Burroughs}, {and} \bibinfo{person}{Clare Jinks}.} \bibinfo{year}{2018}\natexlab{}.
\newblock \showarticletitle{Saturation in qualitative research: exploring its conceptualization and operationalization}.
\newblock \bibinfo{journal}{\emph{Quality \& quantity}}  \bibinfo{volume}{52} (\bibinfo{year}{2018}), \bibinfo{pages}{1893--1907}.
\newblock


\bibitem[Sims et~al\mbox{.}(2020)]%
        {sims2020doing}
\bibfield{author}{\bibinfo{person}{Jennifer~Patrice Sims}, \bibinfo{person}{Whitney~Laster Pirtle}, {and} \bibinfo{person}{Iris Johnson-Arnold}.} \bibinfo{year}{2020}\natexlab{}.
\newblock \showarticletitle{Doing hair, doing race: The influence of hairstyle on racial perception across the US}.
\newblock \bibinfo{journal}{\emph{Ethnic and Racial Studies}} \bibinfo{volume}{43}, \bibinfo{number}{12} (\bibinfo{year}{2020}), \bibinfo{pages}{2099--2119}.
\newblock


\bibitem[Son and Cho(2010)]%
        {son2010study}
\bibfield{author}{\bibinfo{person}{Hyang-Mi Son} {and} \bibinfo{person}{Hyun-Ju Cho}.} \bibinfo{year}{2010}\natexlab{}.
\newblock \showarticletitle{A study on hairstyle in style of subculture}.
\newblock \bibinfo{journal}{\emph{The Research Journal of the Costume Culture}} \bibinfo{volume}{18}, \bibinfo{number}{4} (\bibinfo{year}{2010}), \bibinfo{pages}{755--773}.
\newblock


\bibitem[Tatarian et~al\mbox{.}(2023)]%
        {tatarian2023socioeconomic}
\bibfield{author}{\bibinfo{person}{Talar Tatarian}, \bibinfo{person}{Connor McPartland}, \bibinfo{person}{Lizhou Nie}, \bibinfo{person}{Jie Yang}, \bibinfo{person}{Konstantinos Spaniolas}, \bibinfo{person}{Salvatore Docimo}, {and} \bibinfo{person}{Aurora~D Pryor}.} \bibinfo{year}{2023}\natexlab{}.
\newblock \showarticletitle{Socioeconomic disparities in the utilization of primary robotic hernia repair}.
\newblock \bibinfo{journal}{\emph{Surgical Endoscopy}} \bibinfo{volume}{37}, \bibinfo{number}{6} (\bibinfo{year}{2023}), \bibinfo{pages}{4829--4833}.
\newblock


\bibitem[Ward and Holland(2011)]%
        {ward_if_2011}
\bibfield{author}{\bibinfo{person}{Richard Ward} {and} \bibinfo{person}{Caroline Holland}.} \bibinfo{year}{2011}\natexlab{}.
\newblock \showarticletitle{‘{If} {I} look old, {I} will be treated old’: hair and later-life image dilemmas}.
\newblock \bibinfo{journal}{\emph{Ageing and Society}} \bibinfo{volume}{31}, \bibinfo{number}{2} (\bibinfo{date}{Feb.} \bibinfo{year}{2011}), \bibinfo{pages}{288--307}.
\newblock
\showISSN{0144-686X, 1469-1779}
\urldef\tempurl%
\url{https://doi.org/10.1017/S0144686X10000863}
\showDOI{\tempurl}


\bibitem[Wei et~al\mbox{.}(2021)]%
        {counterfactual_1}
\bibfield{author}{\bibinfo{person}{Tianxin Wei}, \bibinfo{person}{Fuli Feng}, \bibinfo{person}{Jiawei Chen}, \bibinfo{person}{Ziwei Wu}, \bibinfo{person}{Jinfeng Yi}, {and} \bibinfo{person}{Xiangnan He}.} \bibinfo{year}{2021}\natexlab{}.
\newblock \showarticletitle{Model-Agnostic Counterfactual Reasoning for Eliminating Popularity Bias in Recommender System}. In \bibinfo{booktitle}{\emph{Proceedings of the 27th ACM SIGKDD Conference on Knowledge Discovery \& Data Mining}} (Virtual Event, Singapore) \emph{(\bibinfo{series}{KDD '21})}. \bibinfo{publisher}{Association for Computing Machinery}, \bibinfo{address}{New York, NY, USA}, \bibinfo{pages}{1791–1800}.
\newblock
\showISBNx{9781450383325}
\urldef\tempurl%
\url{https://doi.org/10.1145/3447548.3467289}
\showDOI{\tempurl}


\bibitem[Williams(2023)]%
        {williams2023eye}
\bibfield{author}{\bibinfo{person}{Tom Williams}.} \bibinfo{year}{2023}\natexlab{}.
\newblock \showarticletitle{The Eye of the Robot Beholder: Ethical Risks of Representation, Recognition, and Reasoning over Identity Characteristics in Human-Robot Interaction}. In \bibinfo{booktitle}{\emph{Proceedings of the 2023 ACM/IEEE International Conference on Human-Robot Interaction}}.
\newblock


\bibitem[Zhang et~al\mbox{.}(2022)]%
        {counterfactual_2}
\bibfield{author}{\bibinfo{person}{Yi Zhang}, \bibinfo{person}{Junyang Wang}, {and} \bibinfo{person}{Jitao Sang}.} \bibinfo{year}{2022}\natexlab{}.
\newblock \showarticletitle{Counterfactually Measuring and Eliminating Social Bias in Vision-Language Pre-Training Models}. In \bibinfo{booktitle}{\emph{Proceedings of the 30th ACM International Conference on Multimedia}} (<conf-loc>, <city>Lisboa</city>, <country>Portugal</country>, </conf-loc>) \emph{(\bibinfo{series}{MM '22})}. \bibinfo{publisher}{Association for Computing Machinery}, \bibinfo{address}{New York, NY, USA}, \bibinfo{pages}{4996–5004}.
\newblock
\showISBNx{9781450392037}
\urldef\tempurl%
\url{https://doi.org/10.1145/3503161.3548396}
\showDOI{\tempurl}


\end{thebibliography}
\end{document}